\documentclass{article}

\usepackage{PRIMEarxiv}

\usepackage[utf8]{inputenc} % allow utf-8 input
\usepackage[T1]{fontenc}    % use 8-bit T1 fonts
\usepackage{hyperref}       % hyperlinks
\usepackage{url}            % simple URL typesetting
\usepackage{booktabs}       % professional-quality tables
\usepackage{amsfonts}       % blackboard math symbols
\usepackage{nicefrac}       % compact symbols for 1/2, etc.
\usepackage{microtype}      % microtypography
\usepackage{lipsum}
\usepackage{fancyhdr}       % header
\usepackage{graphicx}       % graphics
\graphicspath{{media/}}     % organize your images and other figures under media/ folder

\usepackage{subfigure}
\graphicspath{{figures/}}

\usepackage{xcolor}
\usepackage{amsmath,amssymb,amsfonts,amsthm}
\usepackage{ulem}
\usepackage{enumitem}
\usepackage{mathrsfs}
\usepackage{bbding}
\usepackage{adjustbox}
\usepackage{multirow}
\usepackage{wrapfig}

\usepackage{float}

\RequirePackage{amssymb}

\title{Unveiling and Controlling Anomalous Attention Distribution in Transformers
\thanks{\textit{\underline{Citation}}: 
\textbf{Yan et al. Unveiling and Controlling Anomalous Attention Distribution in Transformers.}} 
}

\author{
  \textbf{Ruiqing Yan\textsuperscript{1}, Xingbo Du\textsuperscript{1,2}, Haoyu Deng\textsuperscript{3,4}, Linghan Zheng\textsuperscript{5}, Qiuzhuang Sun\textsuperscript{6}, Jifang Hu\textsuperscript{3,4},} \\
  \textbf{Yuhang Shao\textsuperscript{4,7}, Penghao Jiang\textsuperscript{9}, Jinrong Jiang\textsuperscript{3,4,8}, Lian Zhao\textsuperscript{3,4}} \\
  \textsuperscript{1}School of Computer Science and Engineering, University of New South Wales, Sydney, Australia \\
  \texttt{\{ruiqing.yan, xingbo.du\}@unsw.edu.au} \\
  \textsuperscript{2}School of Electronic Information and Electrical Engineering, Shanghai Jiao Tong University, Shanghai, China \\
  \texttt{duxingbo@sjtu.edu.cn} \\
  \textsuperscript{3}Computer Network Information Center, Chinese Academy of Sciences, Beijing, China \\
  \textsuperscript{4}University of Chinese Academy of Sciences, Beijing, China \\
  \texttt{\{hydeng, jfhu\}@cnic.cn \{jjr, zhaolian\}@sccas.cn} \\
  \textsuperscript{5}Ant Group, Shanghai, China \\
  \texttt{zhenglinghan.zlh@antgroup.com} \\
  \textsuperscript{6}School of Mathematics and Statistics, The University of Sydney, Sydney, Australia \\
  \texttt{qiuzhuang.sun@sydney.edu.au} \\
  \textsuperscript{7}International Center for Climate and Environment Sciences, \\
  Institute of Atmospheric Physics, Chinese Academy of Sciences, Beijing, China \\
  \texttt{shaoyuhang@mail.iap.ac.cn} \\
  \textsuperscript{8}Hangzhou Institute for Advanced Study, UCAS, Hangzhou, China \\
  \textsuperscript{9}Beijing Xing Rui Qi Zhi Technology Co., Ltd., Beijing, China
}

\begin{document}
\maketitle

\begin{abstract}
% Background
With the advent of large models based on the Transformer architecture, researchers have observed an anomalous phenomenon in the Attention mechanism---there is a very high attention on the first element, which is prevalent across Transformer-based models.
% Motivation
It is crucial to understand it for the development of techniques focusing on attention distribution, such as Key-Value (KV) Cache compression and infinite extrapolation; however, the latent cause leaves to be unknown. 
% Main Contribution
In this paper, we analyze such a phenomenon from the perspective of \textit{waiver} phenomenon, which involves reducing the internal values of certain elements in the sequence, allowing them to absorb excess attention without affecting their contribution to information.
% Detail the Contribution
In specific models, due to differences in positional encoding and attention patterns, we have found that the selection of \textit{waiver} elements by the model can be categorized into two methods: \textbf{positional-encoding-based} and \textbf{feature-distribution-within-elements-based}.
% Experiment results

% Depending on the structured mask matrices, attention can be divided into different variants including global and causal attention. Based on our observations, we hypothesize that the anomalously high focus on the first element in both global and causal attention mechanisms is a ``waiver phenomenon'' learned by the model under Softmax constraints. This phenomenon involves reducing the internal values of certain elements, allowing them to absorb excess attention without affecting their contribution to information. In specific models, due to differences in positional encoding and attention patterns, we have found that the selection of ``waiver'' elements by the model can be categorized into two methods: ``positional-encoding-based'' and ``feature-distribution-within-elements-based''.
\end{abstract}

% keywords can be removed

% \begin{figure}[tb!]
% \centering
% \includegraphics{test}
% \caption{\label{fig:xxx} }
% \end{figure}

\section{Introduction}
% Background
Transformer architectures~\cite{vaswani2017attention,Devlin2018BERT,radford2018improving} have become the most popular foundational structures in deep learning due to their remarkable ability to effectively model long-range dependencies~\cite{lin2022survey} and their scalable parameterization~\cite{tay2022scaling}, enabling excellent learning capabilities. At the core of these models is the attention mechanism, which dynamically assigns weights to input elements, allowing the model to effectively prioritize relevant information. The impact and significance of Transformer models in the field of artificial intelligence are profound, as they have revolutionized natural language processing, computer vision, and other AI applications, establishing themselves as a cornerstone of modern AI advancements.

Regardless of its significant impact, an interesting anomaly~\cite{zaheer2020bigbird,xiao2024efficient,fu2024attentionpattern} has been observed in the attention mechanisms~\cite{vaswani2017attention} of Transformer models: a disproportionate emphasis on the first element in a sequence. This counterintuitive behavior is widespread across multiple Transformer-based architectures~\cite{ge2024model}, necessitating a deeper investigation into its origins and consequences.

It is quite vital to understand such anomaly, because: 1) it impacts the efficiency of KV cache compression~\cite{ge2024model}, a crucial technique for deploying large models in resource-constrained environments. Existing studies~\cite{ge2024model} have leveraged attention distribution patterns to develop adaptive KV cache compression methods, significantly reducing memory usage during generative inference with minimal impact on performance \cite{ge2024model}; 2) this phenomenon affects methods that rely on attention distribution patterns, such as infinite extrapolation, making it essential to clarify the underlying factors driving this behavior; 3) recent research~\cite{xiao2024efficient} has highlighted the importance of attention sinks, where initial tokens act as a repository for excess attention weights. This mechanism helps maintain performance in streaming applications despite finite attention windows \cite{xiao2024efficient}. Moreover, they have utilized attention sinks to achieve infinite extrapolation, demonstrating the critical role of understanding attention mechanisms to enhance the efficiency and scalability of Transformer models.

Although existing research \cite{ge2024model, xiao2024efficient} has identified the tendency of attention mechanisms in Transformers to focus significantly on the first element and has utilized these patterns to improve Transformer performance, a comprehensive explanation remains elusive. This paper aims to delve into such phenomenon under the perspective of \textit{waiver}, which means that the model reduces the weights of certain elements in the Softmax function to absorb excess attention without significantly impacting their interaction with other elements. Building on this, we investigated how trained natural language models select \textit{waiver} elements, proposing two selection strategies: \textbf{positional-encoding-based} and \textbf{feature-distribution-within-elements-based}. Additionally, to validate our hypothesis, we designed methods that can arbitrarily control whether an element becomes a waiver element based on the mechanisms of these two selection strategies, achieving significant results in our experiments.

\textbf{The main contributions of this paper are highlighted as follows}:
\begin{itemize}[leftmargin=10pt, topsep=0pt, itemsep=1pt, partopsep=1pt, parsep=1pt]
    \item Unlike previous studies that directly leverage the excessive focus on the first element in Transformer models to enhance performance, we are the first to attempt to explain this phenomenon and propose a coherent explanation from the perspective of the \textit{waiver} phenomenon.
    \item We study the selection process of the element that absorbs excessive attention and identify two methods to regulate whether an element becomes a waiver option based on the feature distribution within the position encoding embedding vector or the element embedding vector.
\end{itemize}

\section{Related Work}
\subsection{Transformers}
The Transformer~\cite{vaswani2017attention} model has rapidly become the prevailing architecture in natural language processing. Its innovative incorporation of the self-attention mechanism significantly enhances performance in sequence-to-sequence tasks. Subsequent innovations have led to widespread adoption and modification of the model, resulting in numerous variants, primarily categorized into BERT~\cite{Devlin2018BERT} (Global Attention) and GPT~\cite{radford2018improving} (Causal Attention) branches. 

Advancements within the global attention branch have led to several significant innovations. RoBERTa~\cite{liu2019roberta} enhances this model's capabilities with longer training sequences; ALBERT~\cite{lan2019albert} contributes further by implementing a parameter-sharing mechanism to boost processing efficiency. The T5~\cite{raffel2020exploring} model broadens the architecture's applicability; XLNet~\cite{yang2019xlnet} advances these methodologies by integrating a generalized autoregressive pretraining approach. These developments showcase the continuous evolution and enrichment of global attention architectures in natural language processing. 

The causal attention branch of Transformer models also significantly impacts natural language processing, particularly celebrated for its implementations of large-scale language models. Major developments within this lineage include OpenAI's GPT series~\cite{radford2019language,brown2020language,ouyang2022training,achiam2023gpt},  Meta's LLaMA series~\cite{touvron2023llama,touvron2023llama2,meta2024llama3}, Google's PaLM series~\cite{chowdhery2023palm,anil2023palm}. These series collectively emphasize the transformative role of causal attention architectures in advancing the complexities of language understanding, generation, and the broad application of AI in everyday technology. 

Given the substantial computational demands of the Transformer model, enhancing model efficiency while maintaining performance has emerged as a critical area of focus within the community. Researchers have proposed numerous effective improvements, including the Transformer-XL~\cite{dai2019transformer},  Linformer~\cite{wang2020linformer}, Adaptive Sparse Transformer~\cite{correia2019adaptively}, Reformer~\cite{kitaev2020reformer}. Besides the work above, many other ongoing improvements in various aspects of Transformer models exist. However, all these efforts underscore the importance of a deep understanding of the intrinsic mechanisms of Transformer models for driving advancements.

\subsection{Attention Sinks}
As early as 2020, Manzil Zaheer et al.~\cite{zaheer2020bigbird} discovered that transformer-based natural language models tend to focus more on the first few elements in a sequence. Based on this observation, they designed the \textit{Big Bird} model, which calculates attention only on specific elements \cite{zaheer2020bigbird}. 

In late 2022, with the release of \textit{ChatGPT-3.5} by OpenAI, researchers began to focus on large models. Large models have strong learning capabilities, enabling them to generate very long outputs, which poses significant challenges to hardware systems' caching and computational capabilities. To address these challenges, Xiao et al.~\cite{xiao2024efficient} designed an efficient inference method that retains the first few tokens, achieving infinite extrapolation in large model scenarios. In their work, they pointed out that the first four elements absorb a lot of attention, referring to this phenomenon as \textbf{Attention sink} \cite{xiao2024efficient}. Therefore, excluding the first few elements can cause severe numerical instability in attention distribution, preventing the model from functioning properly. 

In 2024, Yao Fu~\cite{fu2024attentionpattern} analyzed the distribution in the \textit{LLaMA-2-7B-80K}~\cite{touvron2023llama2} model and found that most layers in the model allocate more attention to the first few elements. This phenomenon provided new insights for other researchers to compress the KV Cache. Subsequently, Suyu Ge et al.~\cite{ge2024model} designed a KV Cache compression strategy called \textit{FastGen} based on the attention distribution, effectively improving the inference performance of various transformer-based natural language models \cite{ge2024adaptive}.

\section{Analysis from a \textit{Waiver} Perspective}
\subsection{\textit{Waiver} and its Latent Cause}
Self-attention~\cite{vaswani2017attention} plays a pivotal role in the contemporary machine learning communities, with the formulation

\begin{equation}
\text{Attention}(Q, K, V) = \text{softmax}\left(\frac{QK^\top}{\sqrt{d_k}}\right)V, 
\label{eq1}
\end{equation}
where $Q$, $K$, and $V$ are the Query, Key, and Value matrices, respectively, and $d_k$ is the dimension of the key vectors. The calculation of each attention weights $a(i,j)$ by the softmax function is:

\begin{equation}
a(i,j) = \frac{\exp(e(i,j))}{\sum_{k=0}^{seqlen} \exp(e(i,k))}, 
\label{eq2}
\end{equation}
where $e(i,j)$ represents the input of the softmax function, $i$ and $j$ indicate positions in the attention matrix $\mathbb{R}^{B \times \text{num\_heads} \times L \times L}$, where $i$ corresponds to the first $L$ (the row index) and $j$ corresponds to the second $L$ (the column index), and $seqlen$ represents the length of the input sequence. Note that softmax forces the sum of the output to be 1. Now, let us assume an extreme scenario where we focus on the attention of the last element on all elements. In the multi-head self-attention scenario, a situation may arise where the last element within the current head has little to no relationship with each element. However, under the constraint of the softmax output summing to 1, some attention must still be allocated to certain elements. Therefore, the least impactful strategy is to choose an element as a waiver option, so that even if it is allocated a large amount of attention, it will not significantly affect the final weighted sum.

\subsection{Properties of Waiver Elements}
The self-attention mechanism essentially involves computing attention weights and performing a weighted sum with the Value ($V$) matrix to mix the elements in the sequence. Specifically, the attention weights are calculated by multiplying the transposed Query ($Q$) and Key ($K$) matrices. These weights are then used to perform a weighted sum with the $V$ matrix, facilitating token-to-token information interaction. Each individual weighted operation can be expressed as:

\begin{equation}
\text{outs}(i) = \sum_{j=0}^{seqlen} a(i,j) \cdot v(j), 
\label{eq3}
\end{equation}
where $ a(i,j) $ represents the attention weights, $ v(j) $ represents the values in the $V$ matrix, and $seqlen$ represents the length of the input sequence. Note that each term in the summation of Eq.~\eqref{eq3} is a scalar multiplied by a vector, where the scalar is the corresponding attention weight and the vector is the corresponding value from the $V$ matrix. For an element to act as a waiver option, its value in the $V$ matrix should be sufficiently small, so that even when multiplied by a large attention weight scalar, it does not exceed the magnitude of other elements' vectors in the $V$ matrix multiplied by their corresponding attention weight scalars. An effective way to measure the magnitude of a vector is based on the L1 or L2 norm. Formally,
\begin{equation}
\|v(j)\|_1 = \sum_{k} |v_k(j)|, \quad \|v(j)\|_2 = \sqrt{\sum_{k} (v_k(j))^2}, 
\label{eq4}
\end{equation}
where $ v_k(j) $ represents the $k$-th component of the vector $ v(j) $.

Since models in the Transformer architecture typically embed elements as relatively dense vectors, we choose the L2 norm to evaluate the magnitude of vectors based on their length in Euclidean space. From Eq.~\eqref{eq4}, it is straightforward to derive Eq.~\eqref{eq5}, which shows that the L2 norm of a scalar multiplied by a vector is equal to the scalar multiplied by the L2 norm of the vector.

\begin{equation}
\|a(i,j) \cdot v(j)\|_2 = a(i,j) \cdot \|v(j)\|_2, 
\label{eq5}
\end{equation}
This indicates that to reduce the magnitude of the corresponding $ v(j) $ multiplied by the attention weight scalar, the L2 norm of the corresponding $ v(j) $ should be reduced. Furthermore, the attention weights are calculated by taking the dot product of the corresponding vectors in $QK$. The dot product calculation can be written as:

\begin{equation}
Q \cdot K_i = \|Q\|_2 \|K_i\|_2 \cos(\theta), 
\label{eq6}
\end{equation}
where $\cos(\theta)$ is the angle between the two vectors.

As mentioned earlier, the waiver strategy is learned by the model when the target element has little relation to most elements. Assuming that most elements and the waiver option element have vectors in the $K$ matrix that do not have components in the same direction as the vector corresponding to the target element in the $Q$ matrix, i.e., $\cos(\theta)$ is less than 0, this means that the vector corresponding to the waiver option element in the $K$ matrix should be "smaller" to yield a relatively large dot product result.

\subsection{How Waiver Elements Work in the Model}
We observed that in the \textit{meta-llama/Meta-Llama-3-8B} (Llama3-8B)~\cite{wolf2020transformers,meta2024llama3} model released by Meta on Huggingface\cite{wolf2020transformers}, when the index of the first token is 128000 (i.e., the starting special token), the model assigns high attention to the first element from the first layer and significantly lowers the L2 norm of the corresponding vector in the $ V $ matrix compared to other elements, indicating a waiver phenomenon.
However, we found that after changing the first token, the model still exhibited the waiver phenomenon on the first element in the third layer. Additionally, in the \textit{google-bert/bert-large-cased-whole-word-masking} (Bert-Large)~\cite{wolf2020transformers} model released by Google on Huggingface, this waiver phenomenon not only appeared on the first element but also on the last element.

We posit that the observed phenomenon is influenced by the attention pattern, which we will delve into within this section. When the first element cannot be distinguished by the attention pattern, the model uses learnable positional encoding to distinguish the waiver option elements. 

\subsubsection{Models Using Causal Attention}
Many popular models, such as the Llama series~\cite{touvron2023llama,touvron2023llama2,meta2024llama3}, use causal attention. The structured mask matrix of causal attention is a lower triangular matrix. The calculation process is
\begin{equation}
\text{CausalAttention}(Q, K, V) = \begin{bmatrix}
1 & 0 & 0 & \cdots & 0 \\
1 & 1 & 0 & \cdots & 0 \\
1 & 1 & 1 & \cdots & 0 \\
\vdots & \vdots & \vdots & \ddots & \vdots \\
1 & 1 & 1 & \cdots & 1 
\end{bmatrix} \odot \text{softmax}\left(\frac{QK^\top}{\sqrt{d_k}}\right)V, 
\label{eq7}
\end{equation}
where the lower triangular matrix represents the causal mask that ensures each position can only attend to previous positions in the sequence, $\odot$ represents the element-wise Hadamard product, $Q$ and $K$ are the Query and Key matrices, $QK^\top$ is their matrix multiplication, $d_k$ is the dimension of the key vectors, and $V$ is the Value matrix.

It maintains an interesting question whether the index of the first token can be \textbf{any value}. Note that Llama3-8B does not apply Rotary Position Embedding (RoPE) \cite{su2021roformer} to the $ V $ matrix. By observing the calculation process of the first element, it can be found that it is not affected by other elements. The first element is processed by the Transformer structure, which is equivalent to being processed by a multi-layer perceptron (MLP). Therefore, in the output of the first layer of the model, it is easy to distinguish the first element by observing each element separately. This is because the other elements are mixed together through the weighted sum of the attention mechanism and multiple projections, and the first element is equivalent to multiple projections of its corresponding vector in the $V$ matrix, as shown in Eq.~\eqref{eq8}. For the first element, Eq.~\eqref{eq8} does not explicitly write out the projections of Q, K, V calculated in the attention mechanism and the projection of the output, because matrix multiplication satisfies the associative law. Therefore, for the calculation of the first element, the projection in the attention mechanism can be moved to the implicit representation in the MLP:
\begin{equation}
\begin{split}
&\text{outs}_{0} = \text{MLP}(\text{emb}_0),  \\
&\text{outs}_{1,2,...,n} = \text{MLP}(\text{Attention}(\text{emb}_0, \text{emb}_1, \ldots, \text{emb}_n)).
\end{split}
\label{eq8}
\end{equation}

% \begin{equation}

% \label{eq10}
% \end{equation}

\begin{figure}[tb!]
\centering
\includegraphics[width=1.0\textwidth]{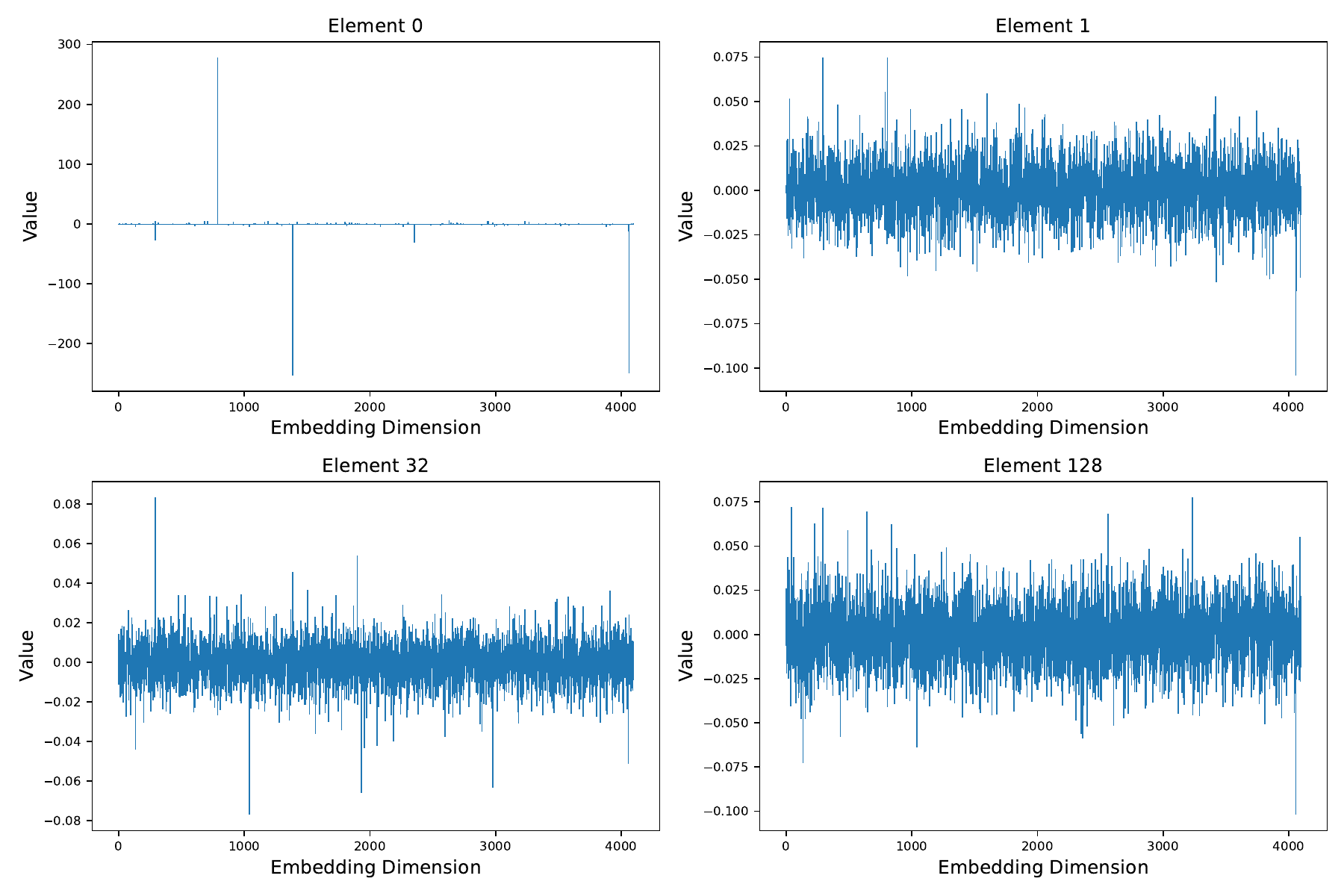}
\caption{The values of the elements of the \textbf{non-mixture distribution} in the FFN output of the second layer of the \textit{Llama3-8B} model.}
\label{fig:distribution}
\end{figure}

With the fixed weights in the MLP, the feature distribution of the first element's vector in the first layer's output should be finite and, more precisely, should be equal to the size of the vocabulary. This distribution should be noticeably different from that of the other elements, which we refer to as \textbf{non-mixed distribution}. Therefore, in theory, all fully connected layers after the first layer of attention mixing can learn to make these \textbf{non-mixed distributions} change differently than other distributions. By observing the attention distribution, we found that excessive attention to the first element appears in the third layer, so we believe that this fully connected layer that can distinguish the first and last elements first appeared in the second layer's feed-forward network (FFN). We observe that in the output of the second layer FFN of the Llama-3-8B model, the vector corresponding to the \textbf{non-mixed distribution} becomes unique, its L2 norm becomes very large, while the ratio of its L1 norm to L2 norm decreases, as shown in Figure~\ref{fig:distribution}, which is significantly different from the characteristic distribution of the vectors corresponding to other elements.

Clearly, starting from the second layer, the model effectively distinguishes the first element from the other elements, creating a significant difference in the distribution of the first element compared to the other elements. In each subsequent layer, the model can distinguish the first element from the other elements based on the different distributions and reduce the L2 norm of the corresponding vectors in the $V$ matrix for specific distributions after projection. We refer to this method of selecting waiver elements as \textbf{feature-distribution-within-elements-based}.

Furthermore, if we generalize to the scenario where the first $ n $ elements of the model input are fixed, the feature distribution of the vectors corresponding to the first $ n $ elements in each layer's hidden states should also be fixed. The model might assign special attention distributions to these fixed distributions or make these fixed features exhibit unique distributions. Specifically, this scenario applies to models that have undergone instruction tuning, which can explain why instruction-tuned models fail severely when the prompt is changed.

\subsubsection{Models Using Global Attention and Learnable Positional Encoding}
In models using global attention and learnable positional encoding, the method for selecting waiver elements can differ significantly. Global attention allows each element in the sequence to attend to every other element, without the constraints imposed by causal attention. This means that each token can be influenced by all other tokens, leading to more complex attention patterns.

We observe that in models using global attention, there is generally higher attention given to the first element, and similarly high attention to the last element. Additionally, we found that the L2 norms of the vectors corresponding to the first and last elements in the $ V $ matrix are also lower than the L2 norms of other elements, indicating that they can serve as waiver options. However, in global attention models, each element is mixed with other elements through the attention mechanism, making it impossible to use the \textbf{based on the feature distribution within elements} method to identify waiver elements.

\begin{figure}[tb!]
\centering
\includegraphics[width=0.8\textwidth]{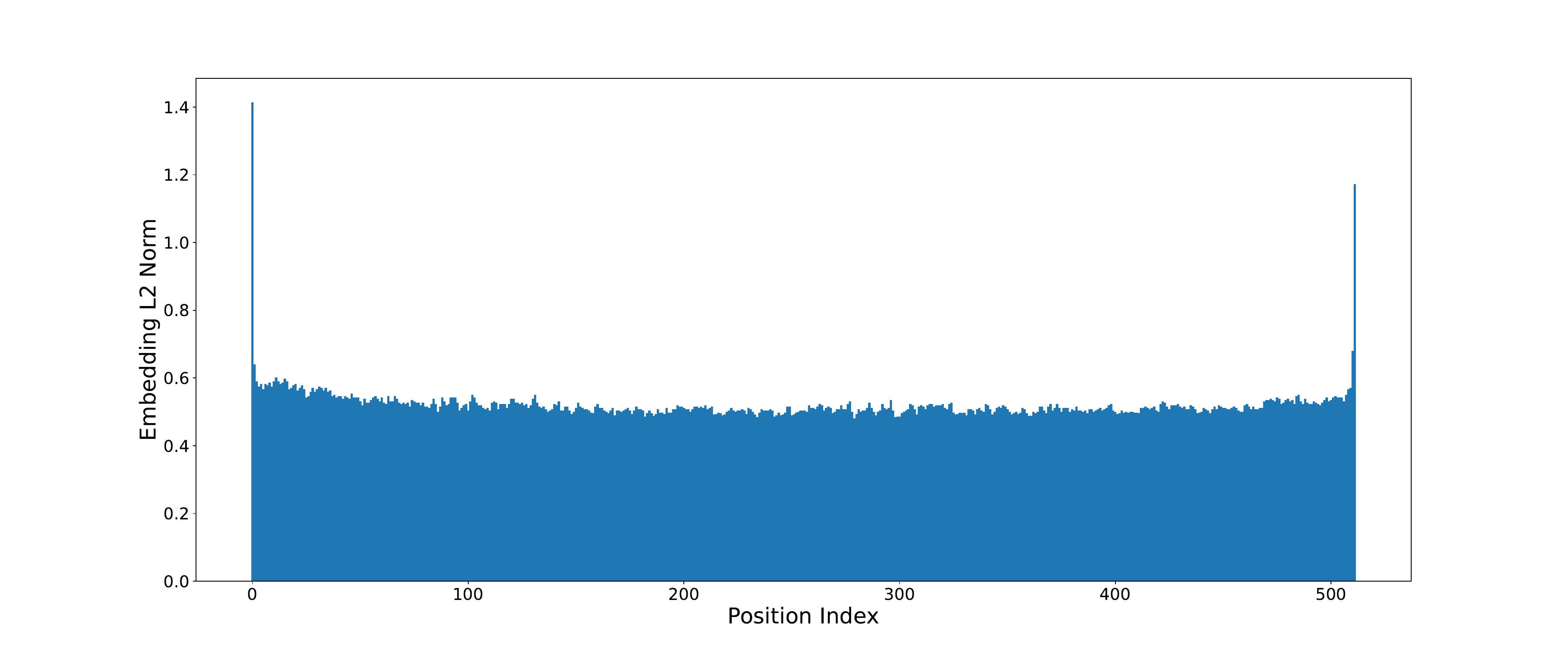}
\caption{The L2 norms of the embedding vectors in the learnable positional encoding of the \textit{Bert-Large} model.}
\label{fig:positional_encoding}
\end{figure}

After examining the pre-trained weights of the \textit{Bert-Large} model, we found that the first and last indices in the learnable positional encoding are assigned embedding vectors with significantly larger L2 norms than other indices, as shown in Figure~\ref{fig:positional_encoding}. Learnable positional encoding is mixed with word embeddings by adding them positionally. This means that elements with the first and last positional indices are assigned biases different from other elements, which might lead to feature distributions distinct from other elements. This distributional difference allows the model to distinguish waiver option elements from other elements.

The reason for the significantly larger L2 norms of the embedding vectors for the first and last positional indices in learnable positional encoding might be that the first and last elements are always \textit{[CLS]} and \textit{[SEP]} during the training of  \textit{Bert-Large} model. Therefore, the model learns these positional encoding embedding vectors differently from other positions. Moreover, we observed that replacing the first element with something other than \textit{[CLS]} and the last element with something other than \textit{[SEP]} results in an increase in the L2 norms of the vectors corresponding to the first and last elements in the $ V $ matrix across layers of the model. However, due to the differing influences of positional encoding embedding vectors, their L2 norms still differ from other elements but not significantly.

Since global attention models rely on the feature distribution of positional encoding embedding vectors to distinguish waiver option elements from other elements, and these positional encoding embedding vectors are fused with token embedding vectors and token type embedding vectors by positional addition before entering the first layer of the Transformer, with no further fusion operations in subsequent layers, the attention mechanism tends to focus more on the first and last elements as early as the first layer.

We refer to this strategy of distinguishing waiver option elements from other elements based on positional encoding as \textbf{positional-encoding-based}. This strategy requires the model to also learn the feature distributions of the first and last elements, as positional information alone is insufficient to effectively distinguish waiver option elements from other elements. 

\section{Experiment}
\subsection{Protocols}
\textbf{Data:} The experimental data is randomly selected from the dataset \textit{bigscience-data/roots\_en\_wikipedia}~\cite{wolf2020transformers}, which consists of English Wikipedia articles. This dataset contains 2.06M samples; considering the visualization burden, we randomly selected 100 samples with a length of over 1024 characters (using Python's built-in len method) for the actual experiments.

\textbf{Models:} The models used in this experiment are \textit{Llama3-8B} and \textit{Bert-Large}. \textit{Llama3-8B} is one of the leading open-source large language models, utilizing rotary position encoding and causal attention. \textit{Bert-Large} is a classic natural language processing model whose encoder part employs global attention and, in the specific implementation on Huggingface, uses a learnable positional encoding with a maximum length of 512.

\subsection{Experimental Design}
We aim to adjust the feature distribution of element embedding vectors to control whether an element becomes a waiver option. By making these adjustments, we hope to observe the attention weights concentrating on the positions we specify. Since the methods for identifying waiver option elements differ between models based on causal attention and models based on global attention, we design two strategies: \textbf{adjusting the structured mask matrix} and \textbf{adjusting the feature distribution within positional encodings}.

\subsubsection{Adjusting the Structured Mask Matrix}
As discussed earlier in this paper, the model treats elements with embedding vectors exhibiting \textit{non-mixed distribution} as waiver elements. Therefore, we need to adjust whether an element is mixed by the attention weighting mechanism. In causal attention, we can control how the attention weighting mechanism affects a particular element by adjusting the structured mask matrix. For example, the vector corresponding to the first element in the structured mask matrix is shown in Eq.~\eqref{eq11}, while the vector corresponding to the $k$-th element is shown in Eq.~\eqref{eq12}. Obviously, if we modify the vector corresponding to the $k$-th element in the structured mask matrix to the form in Eq.~\eqref{eq13}, this element will no longer be mixed with elements in the sequence through the attention weighting mechanism, thereby exhibiting \textit{non-mixed distribution}, and the model will treat this element as a waiver option. Thus, we devise the following attention mask matrices:
\begin{equation}
\text{Mask}_{1, j} = \begin{cases} 
unmask & \text{if } j \leq 1 \\
mask & \text{if } j > 1 
\end{cases}
\label{eq11}
\end{equation}
\begin{equation}
\text{Mask}_{k, j} = \begin{cases} 
unmask & \text{if } j \leq k \\
mask & \text{if } j > k 
\end{cases}
\label{eq12}
\end{equation}
\begin{equation}
\text{Mask}_{k, j}^{\text{modified}} = \begin{cases} 
unmask & \text{if } j = k \\
mask & \text{if } j \neq k .
\end{cases}
\label{eq13}
\end{equation}

Here, $\text{Mask}_{i, j}$ represents the attention mask matrix, where $i$ denotes the row corresponding to the element position in the sequence, and $j$ denotes the column corresponding to the position being attended to. A sign of $unmask$ indicates that the position $j$ is attended to by position $i$, and $mask$ indicates that the attention score is set to $-\infty$ before applying the softmax function, effectively masking the position $j$ from being attended to by position $i$.

By modifying the attention mask matrix as shown in Eq.~\eqref{eq13}, we can ensure that the $k$-th element is not mixed with other elements in the sequence through the attention weighting mechanism, making it exhibit \textit{non-mixed distribution}, and thereby allowing the model to treat this element as a waiver option.

\subsubsection{Adjusting the Feature Distribution within Positional Encodings}
This strategy involves replacing the positional encoding embedding vectors corresponding to other positions with those corresponding to the first and last positions. By doing so, the replaced positions will also have the positional bias of the first and last positions. As discussed earlier in this paper, the model will mark elements with such positional bias as waiver option elements. However, since the tokens corresponding to the selected positions are uncertain, even if more attention is allocated to these positions, the L2 norm of the embedding vectors in the matrix corresponding to these positions may not differ significantly from most other elements.

\subsection{Results and Analysis}
\subsubsection{Experiments with Adjusting the Structured Mask Matrix on Llama3-8B}
\begin{figure}[h]
\centering
\addtocounter{subfigure}{0}
\subfigure[Special token; not changed.]
{\includegraphics[width=0.24\textwidth]{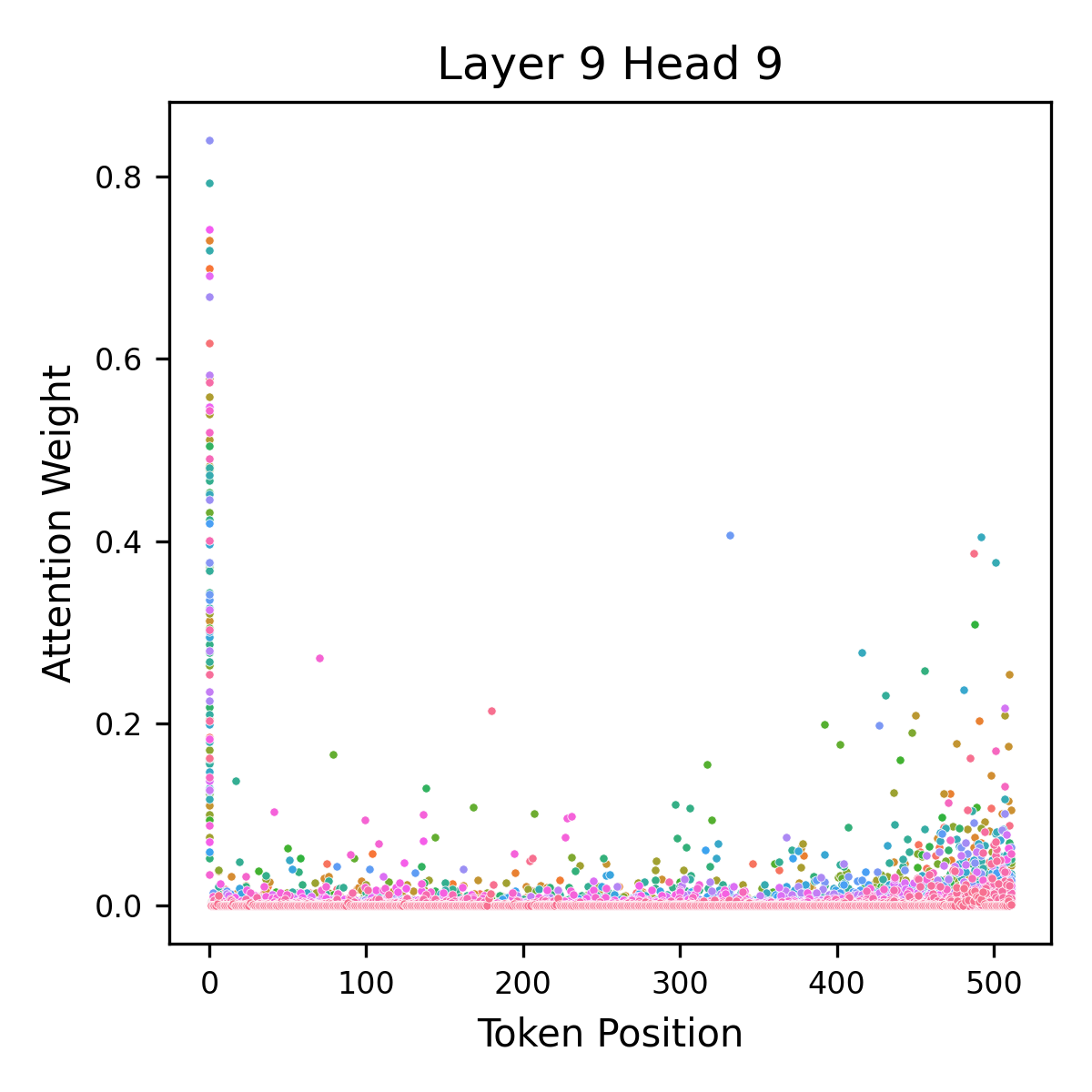}}
\subfigure[Special token; changed.]
{\includegraphics[width=0.24\textwidth]{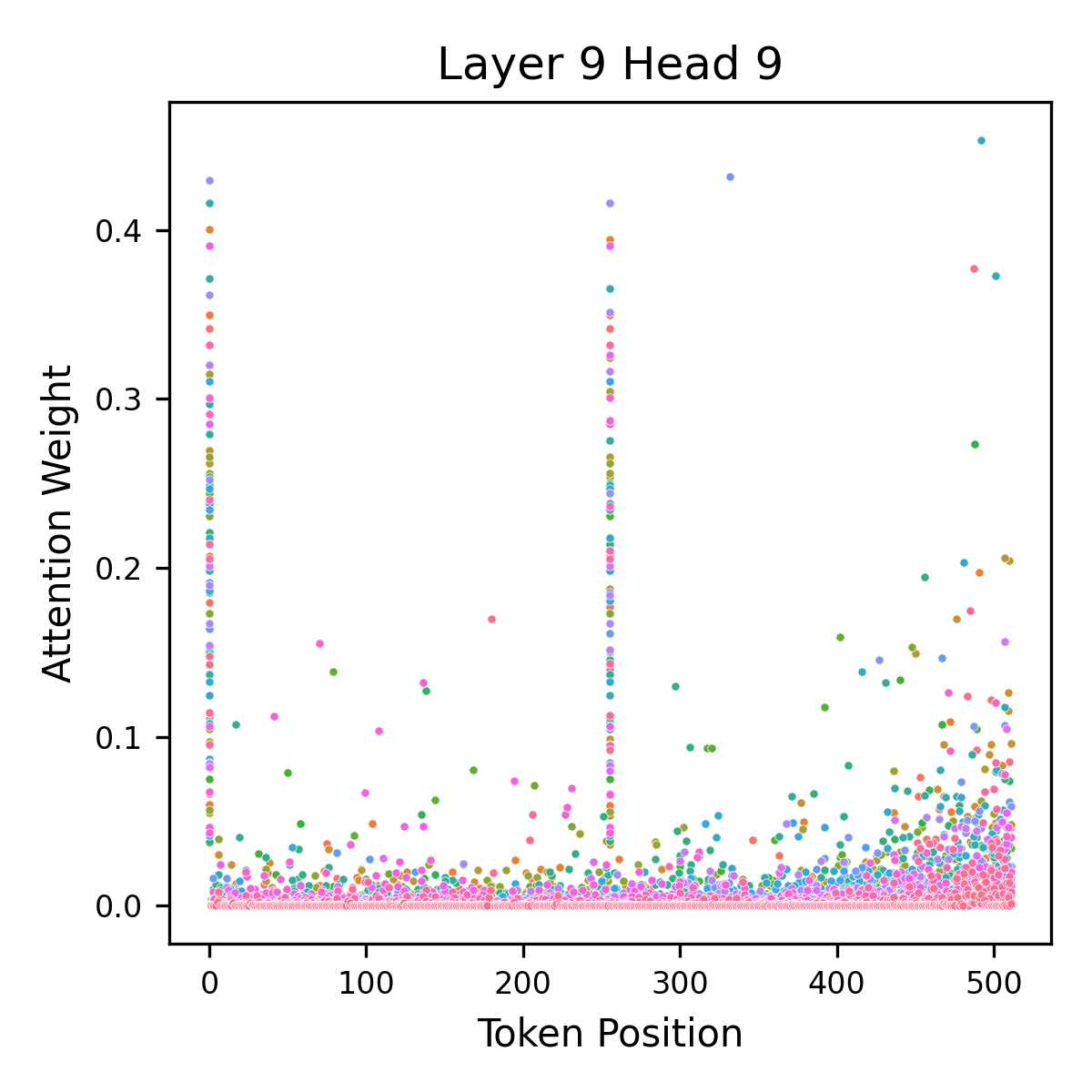}}
\subfigure[Random token; not changed.]
{\includegraphics[width=0.24\textwidth]{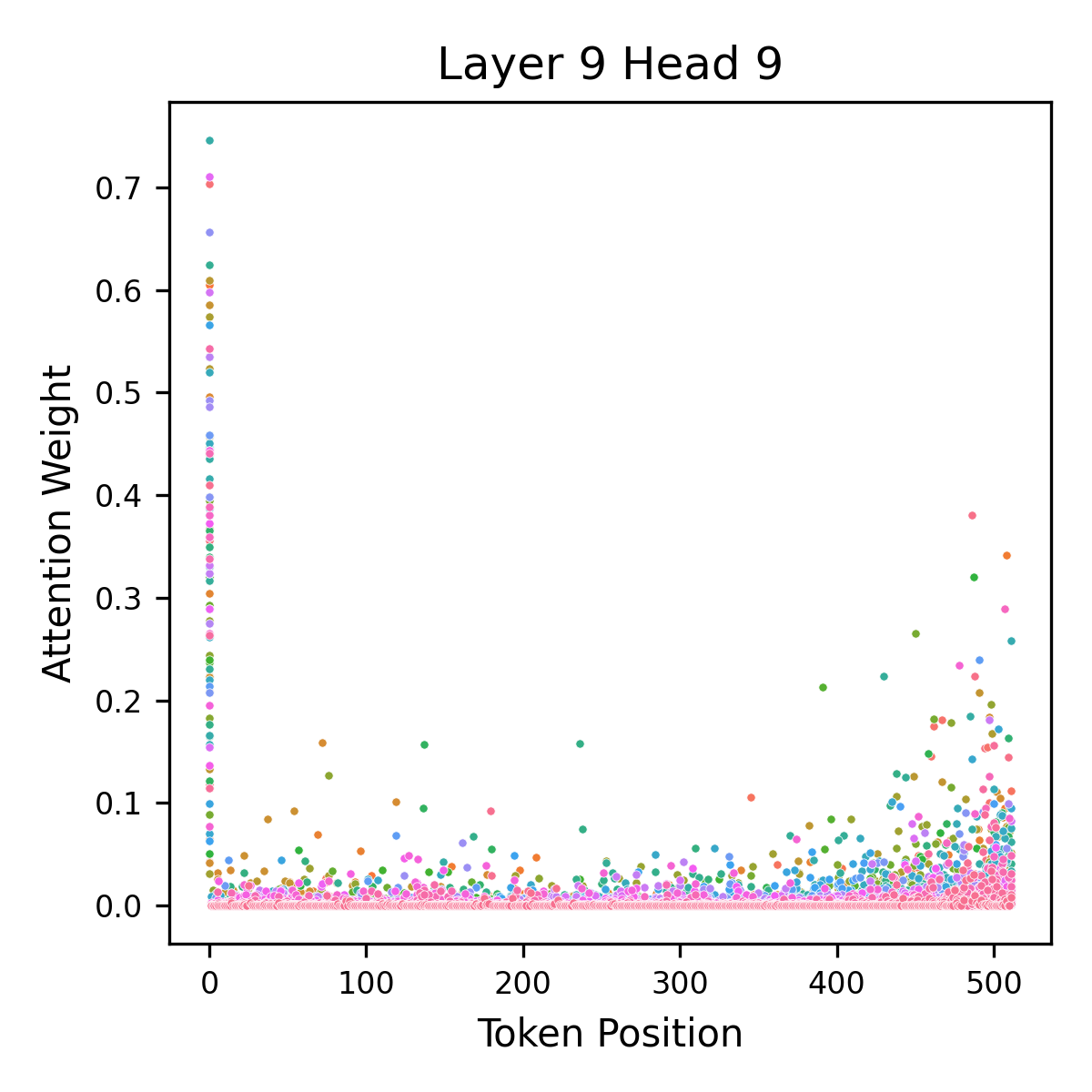}}
\subfigure[Random token; changed.]
{\includegraphics[width=0.24\textwidth]{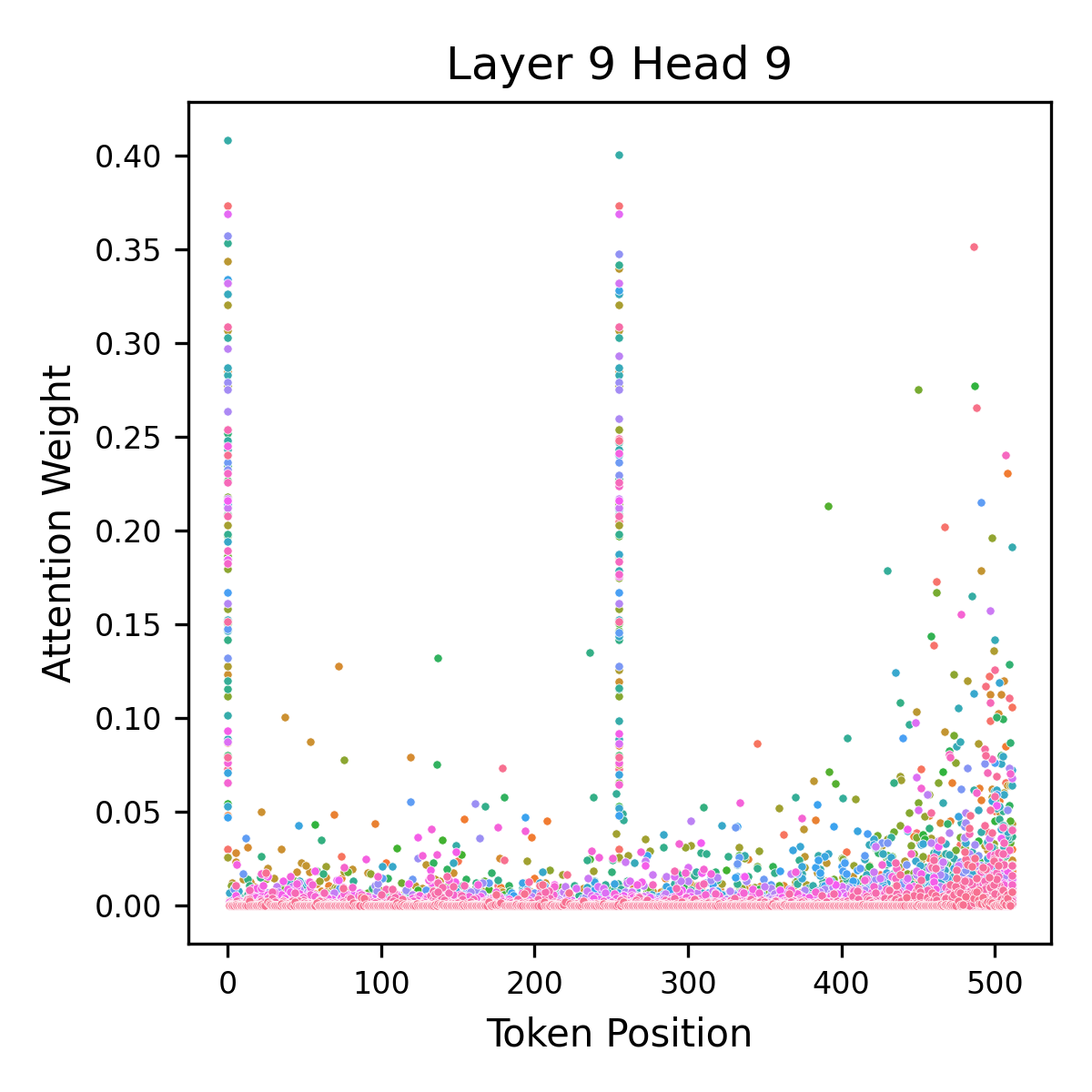}}
\caption{In each subfigure, points of the same color represent elements within a sequence. (a) \textbf{Special} start tokens, the element at position index 255 is \textbf{not changed}; (b) \textbf{Special} start tokens, the element at position index 255 is \textbf{changed}; (c) \textbf{Random} start tokens, the element at position index 255 is \textbf{not changed}; (d) \textbf{Random} start tokens, the element at position index 255 is \textbf{changed}.}
\label{fig:llama_attention}
\end{figure}

We selected the element with the position index \textbf{255} as the artificially designated waiver option, and the element with the position index \textbf{0} corresponding to the word vector index \textbf{128000} and other random values. Since the last element is crucial for generative tasks, we also observed the attention distribution of the last element to the elements in the sequence. Due to the numerous attention heads in the model, it is impractical to display all of them in the paper, so we selected representative results. The experimental results are shown in Figure \ref{fig:llama_attention}.

Clearly, at the adjusted position, there is an attention concentration similar to that of the first element, which aligns with our expectations.

\subsubsection{Experiments with Adjusting the Feature Distribution within Positional Encodings on Bert-Large}
\begin{figure}[h]
\centering
\addtocounter{subfigure}{0}
\subfigure[Special; not changed]
{\includegraphics[width=0.24\textwidth]{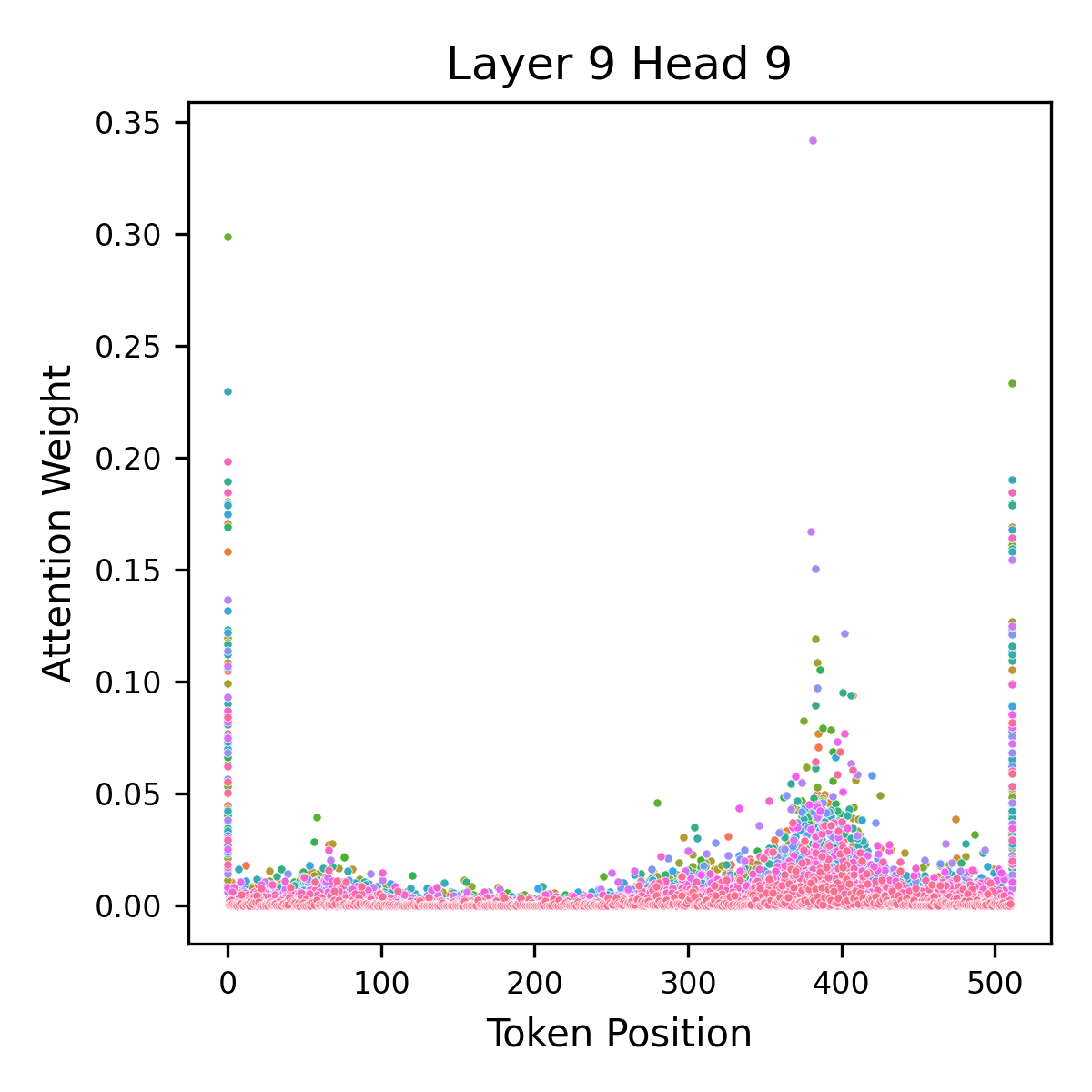}}
\subfigure[Special; changed by first.]
{\includegraphics[width=0.24\textwidth]{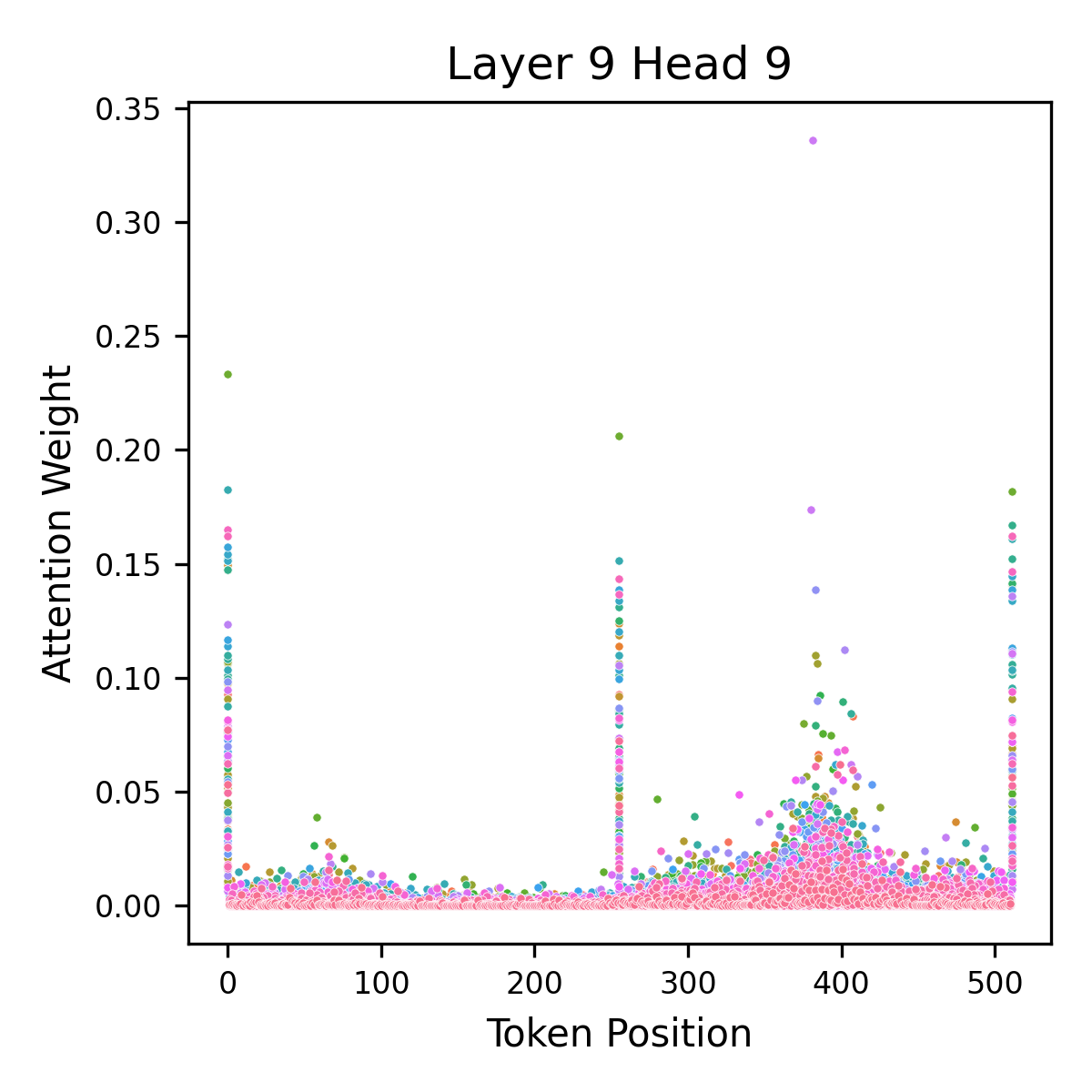}}
\subfigure[Random; not changed.]
{\includegraphics[width=0.24\textwidth]{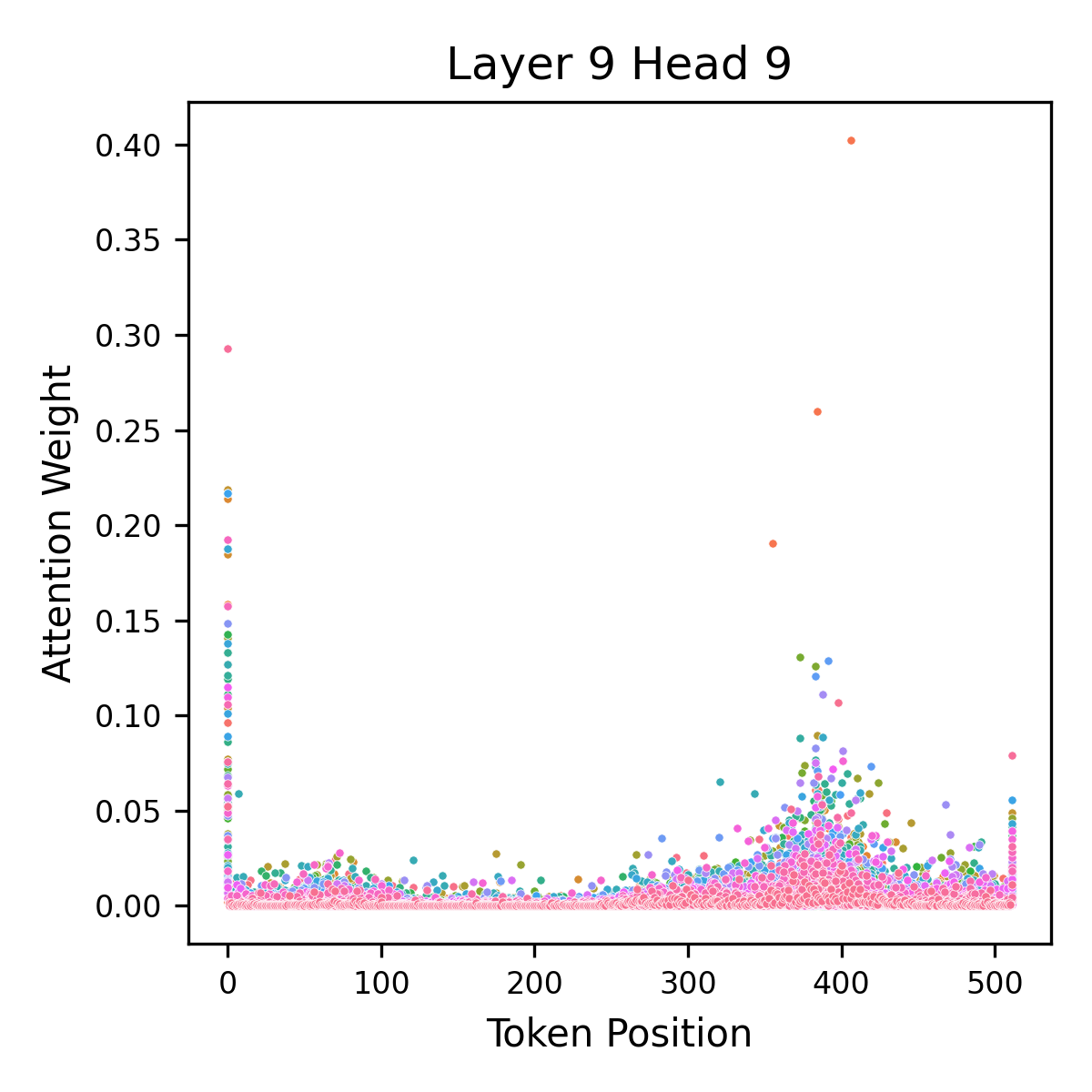}}
\subfigure[Random; changed by first.]
{\includegraphics[width=0.24\textwidth]{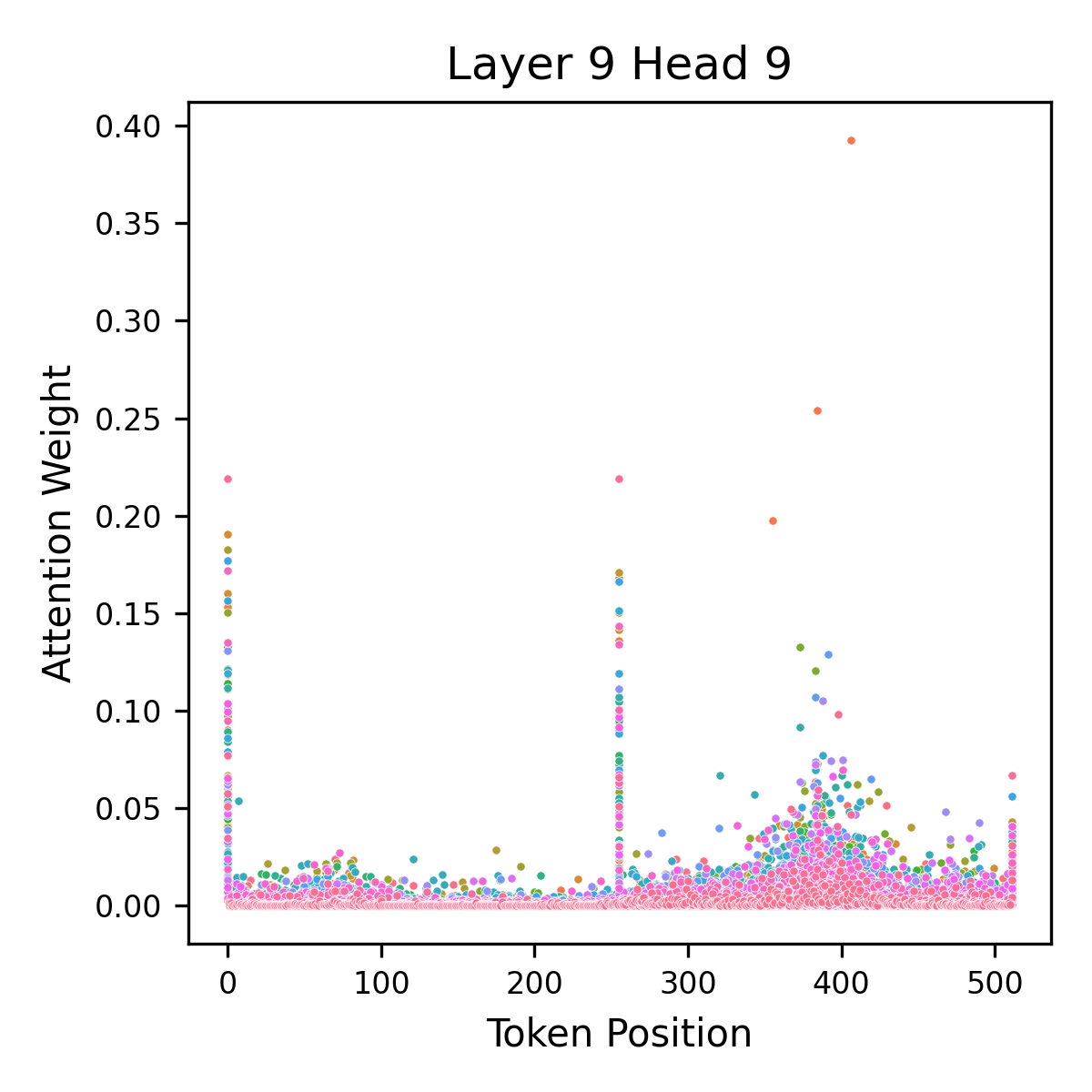}}
\subfigure[Special; not changed.]
{\includegraphics[width=0.24\textwidth]{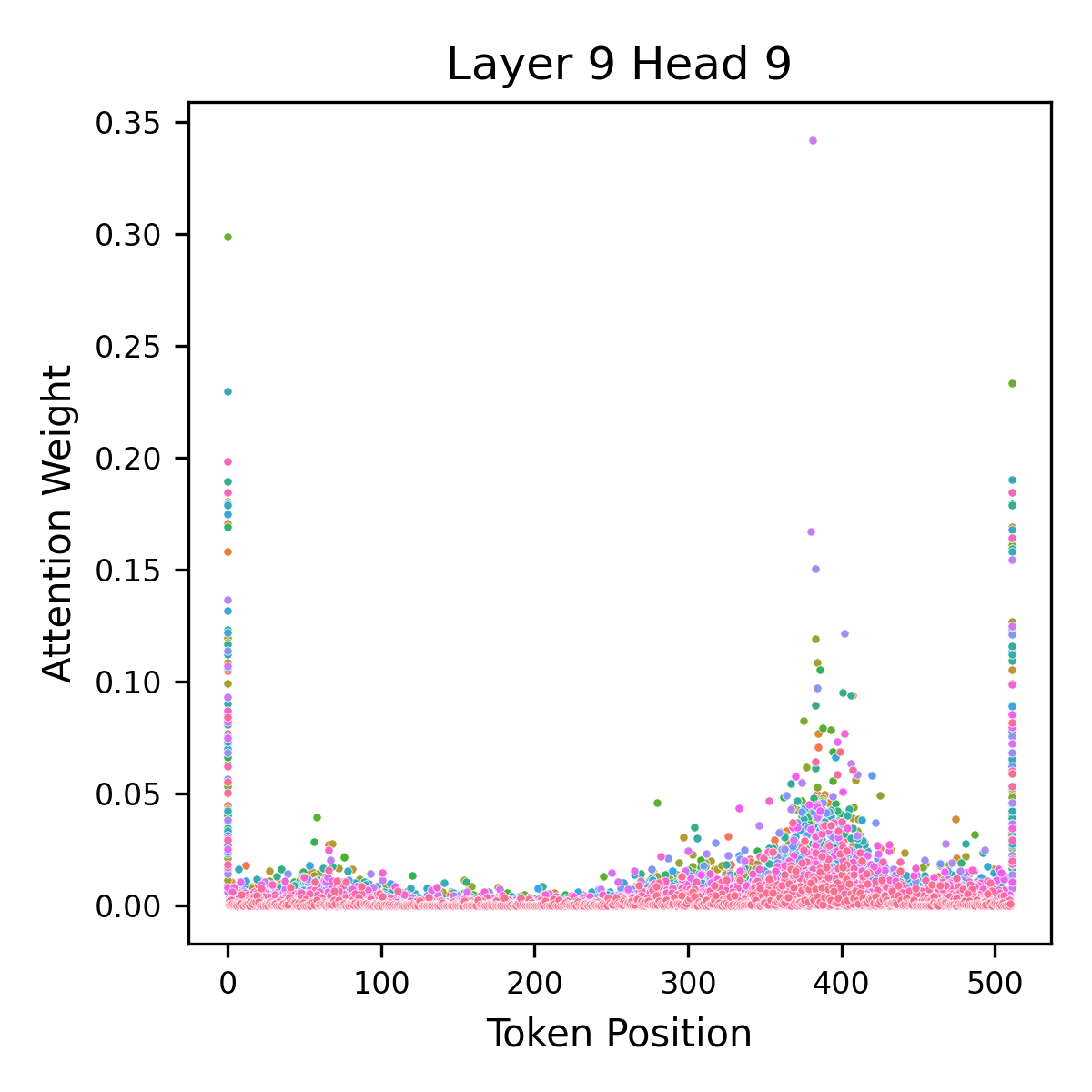}}
\subfigure[Special; changed by last.]
{\includegraphics[width=0.24\textwidth]{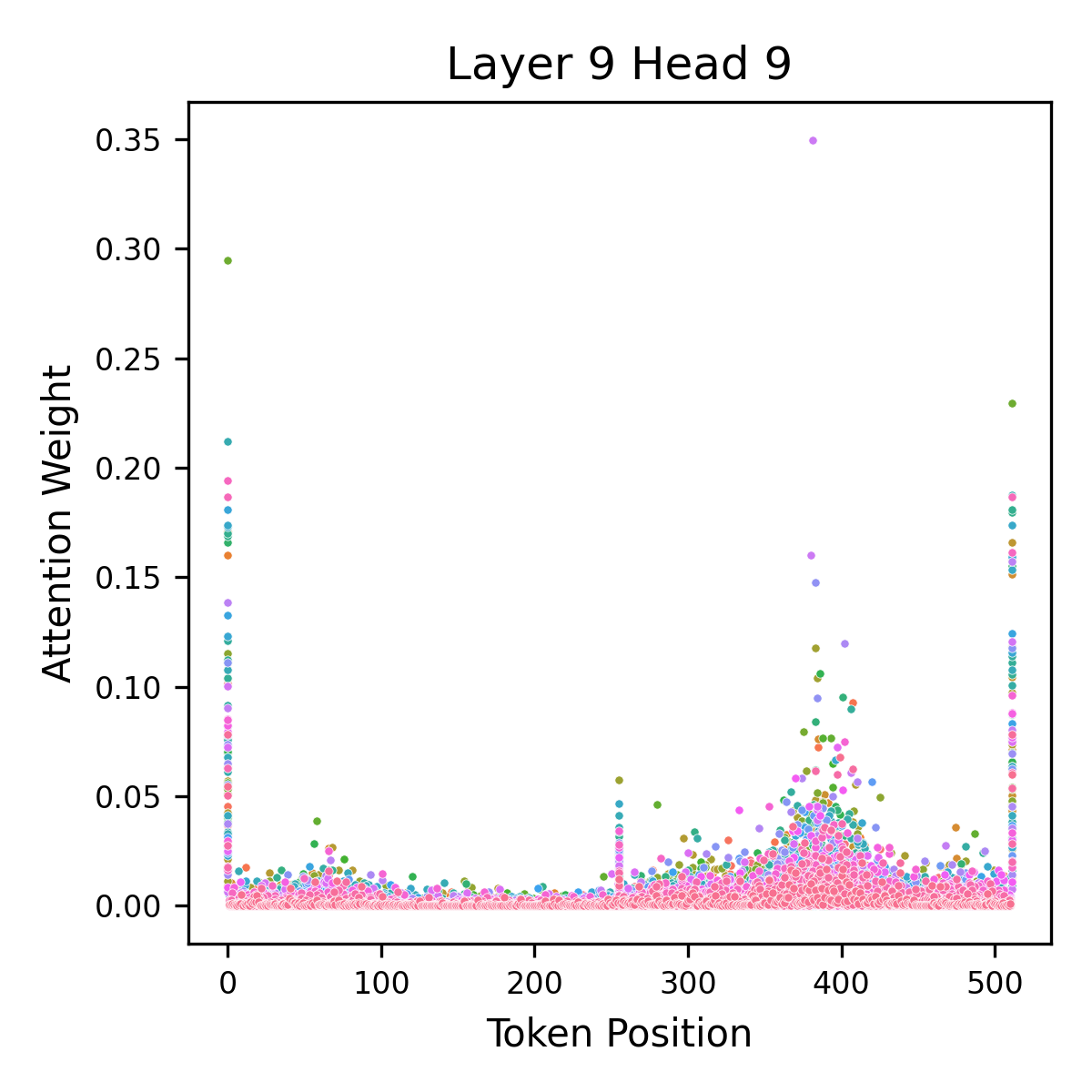}}
\subfigure[Random; not changed.]
{\includegraphics[width=0.24\textwidth]{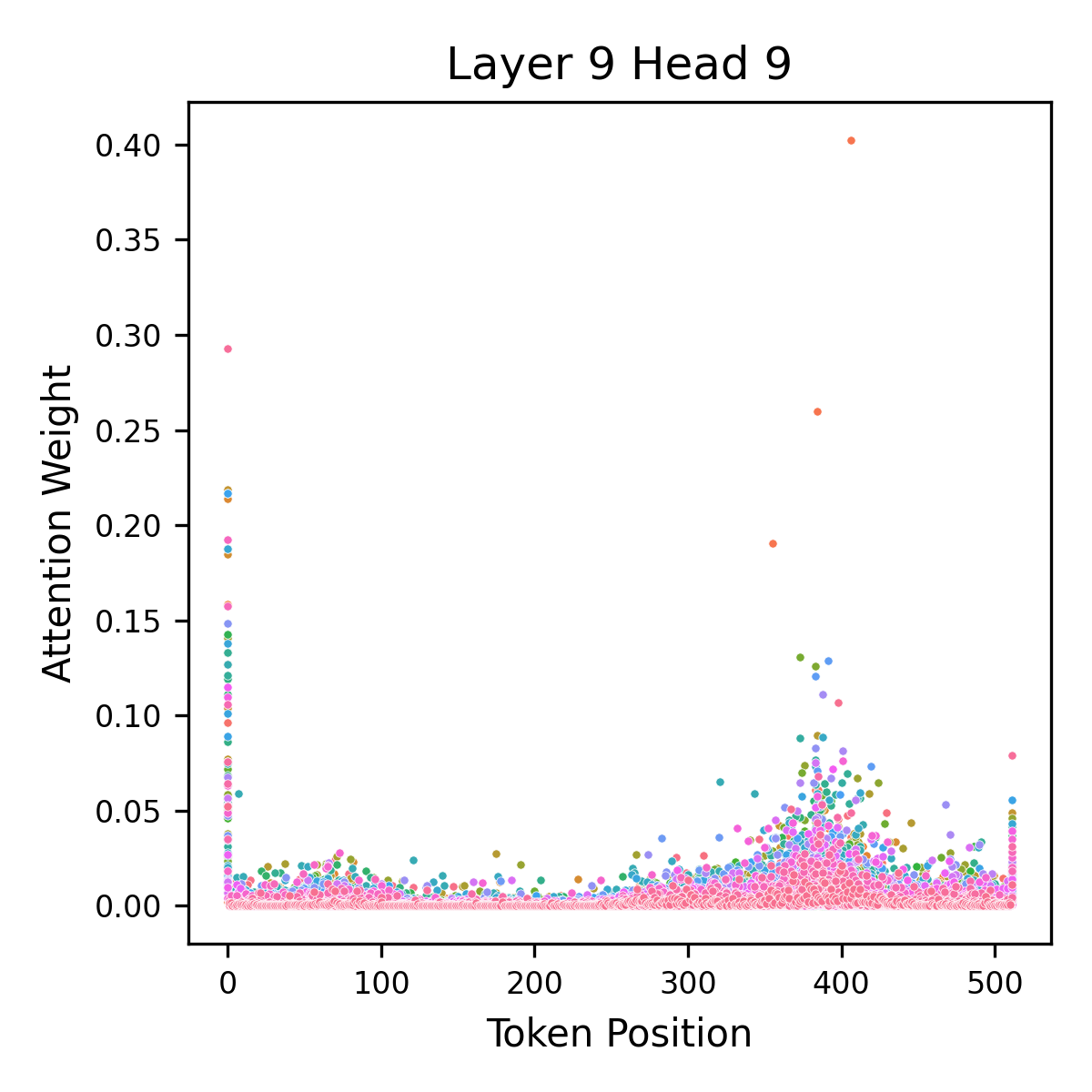}}
\subfigure[Random; changed by last.]
{\includegraphics[width=0.24\textwidth]{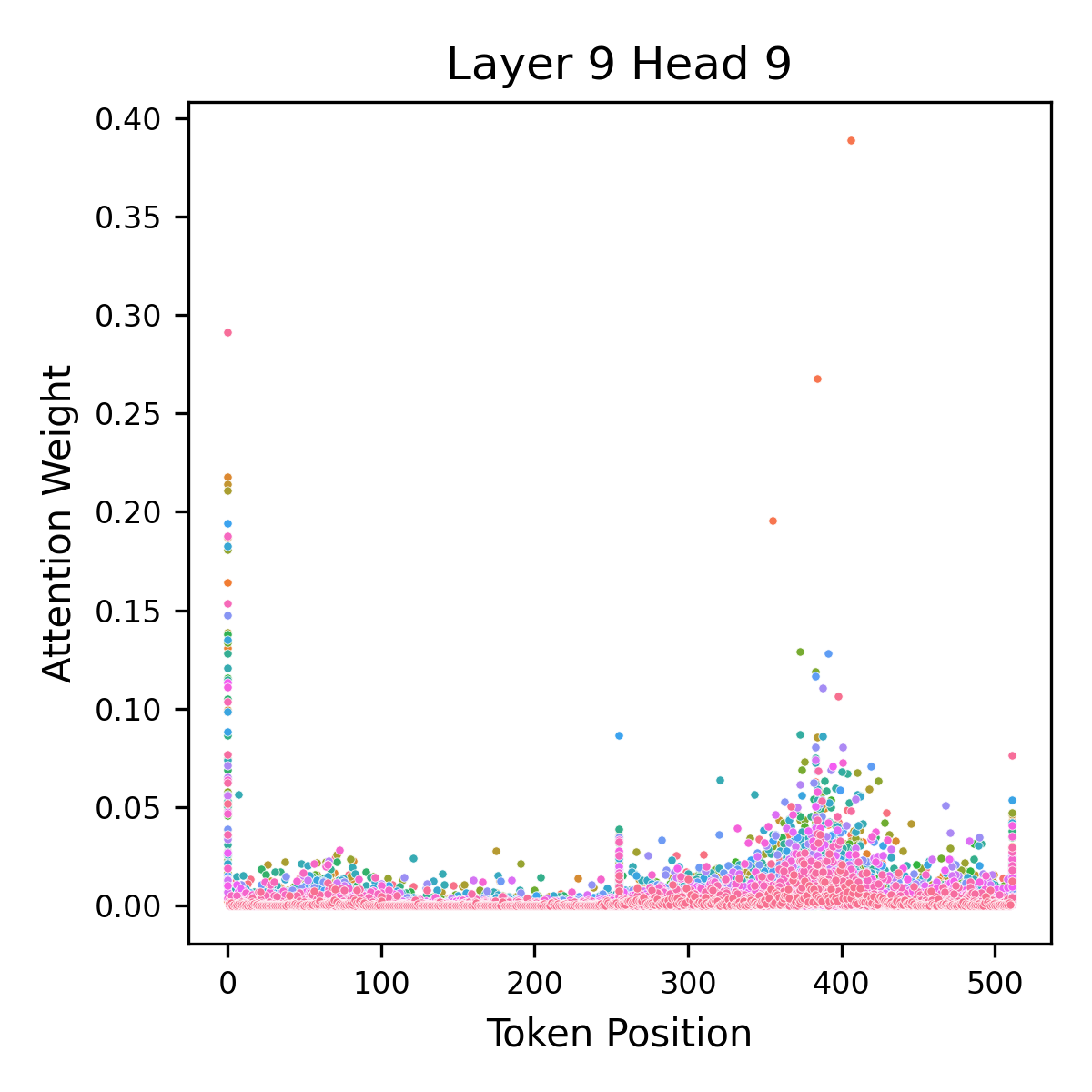}}
\caption{In each subfigure, points of the same color represent elements within a sequence. (a) \textbf{Special} start and end tokens, the element at position index 255 is \textbf{not changed}; (b) \textbf{Special} start and end tokens, the element at position index 255 is \textbf{changed} by \textbf{first} position embedding; (c) \textbf{Random} start and end tokens, the element at position index 255 is \textbf{not changed}; (d) \textbf{Random} start and end tokens, the element at position index 255 is \textbf{changed} by \textbf{first} position embedding; (e) \textbf{Special} start and end tokens, the element at position index 255 is \textbf{not changed}; (f) \textbf{Special} start and end tokens, the element at position index 255 is \textbf{changed} by \textbf{last} position embedding; (g) \textbf{Random} start and end tokens, the element at position index 255 is \textbf{not changed}; (h) \textbf{Random} start and end tokens, the element at position index 255 is \textbf{changed} by \textbf{last} position embedding.}
\label{fig:bert_attention}
\end{figure}

We selected the element with the position index \textbf{383} as the artificially designated waiver option and replaced its positional encoding embedding vector with the vectors corresponding to position indices \textbf{0} and \textbf{511} (the last element) in the positional encoding embedding matrix. Additionally, we set the word vector index of the element at position index \textbf{0} to \textbf{101} and other random values, with a sequence length of 512. To observe the attention weights assigned to the first and last elements, we observed the attention distribution of the element with position index \textbf{383} to the elements in the sequence and selected representative results. The experimental results are shown in Figure \ref{fig:bert_attention}.

At the adjusted position, we observed that the attention given by the model to our adjusted position is similar to the attention given to the first and last elements when their word vector indices are random values. However, there is a certain difference when the word vector indices of the first or last elements are \textbf{101} or \textbf{102}, but both attract more attention compared to other elements. The experimental results also clearly show that the model pays more attention to elements near the position index \textbf{383}, which aligns with linguistic principles that closer elements are more likely to provide relevant information.

\section{Conclusion}
In this paper, we have presented an in-depth analysis of the anomalous phenomenon observed in Transformer models, where a disproportionately high attention is given to the first element in a sequence. We introduced the concept of the \textit{waiver} phenomenon to explain this behavior and proposed two methods for selecting waiver elements: \textbf{positional-encoding-based} and \textbf{feature-distribution-within-elements-based}. 

Through our experiments, we demonstrated that adjusting the structured mask matrix and feature distribution within positional encodings effectively controls whether an element becomes a waiver option. Our findings provide a coherent explanation for the observed attention patterns and offer new insights for improving the efficiency and performance of Transformer-based models.

Future work will focus on further refining these methods and exploring their applications in different Transformer architectures and tasks. We believe that understanding and leveraging the waiver phenomenon can lead to more robust and efficient models, particularly in resource-constrained environments.

\newpage
\section*{Acknowledgments}
This research was supported by the National Key Research and Development Program of China (2022YFE0195900) and the National Key Scientific and Technological Infrastructure project “Earth System Numerical Simulation Facility” (EarthLab).

%Bibliography
\bibliographystyle{unsrt}  
\bibliography{references}  

\begin{thebibliography}{10}

\bibitem{vaswani2017attention}
Ashish Vaswani, Noam Shazeer, Niki Parmar, Jakob Uszkoreit, Llion Jones, Aidan~N Gomez, Lukasz Kaiser, and Illia Polosukhin.
\newblock Attention is all you need.
\newblock {\em arXiv preprint arXiv:1706.03762}, 2017.

\bibitem{Devlin2018BERT}
Jacob Devlin, Ming{-}Wei Chang, Kenton Lee, and Kristina Toutanova.
\newblock {BERT:} pre-training of deep bidirectional transformers for language understanding.
\newblock {\em CoRR}, abs/1810.04805, 2018.

\bibitem{radford2018improving}
Alec Radford, Karthik Narasimhan, Tim Salimans, Ilya Sutskever, et~al.
\newblock Improving language understanding by generative pre-training.
\newblock 2018.

\bibitem{lin2022survey}
Tianyang Lin, Yuxin Wang, Xiangyang Liu, and Xipeng Qiu.
\newblock A survey of transformers.
\newblock {\em AI open}, 3:111--132, 2022.

\bibitem{tay2022scaling}
Yi~Tay, Mostafa Dehghani, Samira Abnar, Hyung~Won Chung, William Fedus, Jinfeng Rao, Sharan Narang, Vinh~Q Tran, Dani Yogatama, and Donald Metzler.
\newblock Scaling laws vs model architectures: How does inductive bias influence scaling?
\newblock {\em arXiv preprint arXiv:2207.10551}, 2022.

\bibitem{zaheer2020bigbird}
Manzil Zaheer, Guru Guruganesh, Avinava Dubey, Joshua Ainslie, Chris Alberti, Santiago Ontanon, Philip Pham, Anirudh Ravula, Qifan Wang, Li~Yang, and Amr Ahmed.
\newblock Big bird: Transformers for longer sequences.
\newblock {\em arXiv preprint arXiv:2007.14062}, 2020.

\bibitem{xiao2024efficient}
Guangxuan Xiao, Yuandong Tian, Beidi Chen, Song Han, and Mike Lewis.
\newblock Efficient streaming language models with attention sinks.
\newblock {\em arXiv preprint arXiv:2309.17453}, 2024.

\bibitem{fu2024attentionpattern}
Yao Fu.
\newblock How do language models put attention weights over long context?, Mar 2024.

\bibitem{ge2024model}
Suyu Ge, Yunan Zhang, Liyuan Liu, Minjia Zhang, Jiawei Han, and Jianfeng Gao.
\newblock Model tells you what to discard: Adaptive kv cache compression for llms.
\newblock {\em arXiv preprint arXiv:2310.01801}, 2024.

\bibitem{liu2019roberta}
Yinhan Liu, Myle Ott, Naman Goyal, Jingfei Du, Mandar Joshi, Danqi Chen, Omer Levy, Mike Lewis, Luke Zettlemoyer, and Veselin Stoyanov.
\newblock Roberta: A robustly optimized bert pretraining approach.
\newblock {\em arXiv preprint arXiv:1907.11692}, 2019.

\bibitem{lan2019albert}
Zhenzhong Lan, Mingda Chen, Sebastian Goodman, Kevin Gimpel, Piyush Sharma, and Radu Soricut.
\newblock Albert: A lite bert for self-supervised learning of language representations.
\newblock {\em arXiv preprint arXiv:1909.11942}, 2019.

\bibitem{raffel2020exploring}
Colin Raffel, Noam Shazeer, Adam Roberts, Katherine Lee, Sharan Narang, Michael Matena, Yanqi Zhou, Wei Li, and Peter~J Liu.
\newblock Exploring the limits of transfer learning with a unified text-to-text transformer.
\newblock {\em Journal of machine learning research}, 21(140):1--67, 2020.

\bibitem{yang2019xlnet}
Zhilin Yang, Zihang Dai, Yiming Yang, Jaime Carbonell, Russ~R Salakhutdinov, and Quoc~V Le.
\newblock Xlnet: Generalized autoregressive pretraining for language understanding.
\newblock {\em Advances in neural information processing systems}, 32, 2019.

\bibitem{radford2019language}
Alec Radford, Jeffrey Wu, Rewon Child, David Luan, Dario Amodei, Ilya Sutskever, et~al.
\newblock Language models are unsupervised multitask learners.
\newblock {\em OpenAI blog}, 1(8):9, 2019.

\bibitem{brown2020language}
Tom Brown, Benjamin Mann, Nick Ryder, Melanie Subbiah, Jared~D Kaplan, Prafulla Dhariwal, Arvind Neelakantan, Pranav Shyam, Girish Sastry, Amanda Askell, et~al.
\newblock Language models are few-shot learners.
\newblock {\em Advances in neural information processing systems}, 33:1877--1901, 2020.

\bibitem{ouyang2022training}
Long Ouyang, Jeffrey Wu, Xu~Jiang, Diogo Almeida, Carroll Wainwright, Pamela Mishkin, Chong Zhang, Sandhini Agarwal, Katarina Slama, Alex Ray, et~al.
\newblock Training language models to follow instructions with human feedback.
\newblock {\em Advances in neural information processing systems}, 35:27730--27744, 2022.

\bibitem{achiam2023gpt}
Josh Achiam, Steven Adler, Sandhini Agarwal, Lama Ahmad, Ilge Akkaya, Florencia~Leoni Aleman, Diogo Almeida, Janko Altenschmidt, Sam Altman, Shyamal Anadkat, et~al.
\newblock Gpt-4 technical report.
\newblock {\em arXiv preprint arXiv:2303.08774}, 2023.

\bibitem{touvron2023llama}
Hugo Touvron, Thibaut Lavril, Gautier Izacard, Xavier Martinet, Marie-Anne Lachaux, Timoth{\'e}e Lacroix, Baptiste Rozi{\`e}re, Naman Goyal, Eric Hambro, Faisal Azhar, et~al.
\newblock Llama: Open and efficient foundation language models.
\newblock {\em arXiv preprint arXiv:2302.13971}, 2023.

\bibitem{touvron2023llama2}
Hugo Touvron, Louis Martin, Kevin Stone, Peter Albert, Amjad Almahairi, Yasmine Babaei, Nikolay Bashlykov, Soumya Batra, Prajjwal Bhargava, Shruti Bhosale, et~al.
\newblock Llama 2: Open foundation and fine-tuned chat models.
\newblock {\em arXiv preprint arXiv:2307.09288}, 2023.

\bibitem{meta2024llama3}
{Meta AI}.
\newblock Introducing meta llama 3: The most capable openly available llm to date, 2024.

\bibitem{chowdhery2023palm}
Aakanksha Chowdhery, Sharan Narang, Jacob Devlin, Maarten Bosma, Gaurav Mishra, Adam Roberts, Paul Barham, Hyung~Won Chung, Charles Sutton, Sebastian Gehrmann, et~al.
\newblock Palm: Scaling language modeling with pathways.
\newblock {\em Journal of Machine Learning Research}, 24(240):1--113, 2023.

\bibitem{anil2023palm}
Rohan Anil, Andrew~M Dai, Orhan Firat, Melvin Johnson, Dmitry Lepikhin, Alexandre Passos, Siamak Shakeri, Emanuel Taropa, Paige Bailey, Zhifeng Chen, et~al.
\newblock Palm 2 technical report.
\newblock {\em arXiv preprint arXiv:2305.10403}, 2023.

\bibitem{dai2019transformer}
Zihang Dai, Zhilin Yang, Yiming Yang, Jaime~G Carbonell, Quoc Le, and Ruslan Salakhutdinov.
\newblock Transformer-xl: Attentive language models beyond a fixed-length context.
\newblock In {\em Proceedings of the 57th Annual Meeting of the Association for Computational Linguistics}, pages 2978--2988, 2019.

\bibitem{wang2020linformer}
Sinong Wang, Belinda~Z Li, Madian Khabsa, Han Fang, and Hao Ma.
\newblock Linformer: Self-attention with linear complexity.
\newblock {\em arXiv preprint arXiv:2006.04768}, 2020.

\bibitem{correia2019adaptively}
Gon{\c{c}}alo~M Correia, Vlad Niculae, and Andr{\'e}~FT Martins.
\newblock Adaptively sparse transformers.
\newblock In {\em Proceedings of the 2019 Conference on Empirical Methods in Natural Language Processing and the 9th International Joint Conference on Natural Language Processing (EMNLP-IJCNLP)}, pages 2174--2184, 2019.

\bibitem{kitaev2020reformer}
Nikita Kitaev, {\L}ukasz Kaiser, and Anselm Levskaya.
\newblock Reformer: The efficient transformer.
\newblock {\em arXiv preprint arXiv:2001.04451}, 2020.

\bibitem{ge2024adaptive}
Suyu Ge, Yunan Zhang, Liyuan Liu, Minjia Zhang, Jiawei Han, and Jianfeng Gao.
\newblock Model tells you what to discard: Adaptive kv cache compression for llms.
\newblock {\em arXiv preprint arXiv:2310.01801}, 2024.

\bibitem{wolf2020transformers}
Thomas Wolf, Lysandre Debut, Victor Sanh, Julien Chaumond, Clement Delangue, Anthony Moi, Pierric Cistac, Tim Rault, Remi Louf, Morgan Funtowicz, Joe Davison, Sam Shleifer, Patrick von Platen, Clara Ma, Yacine Jernite, Julien Plu, Canwen Xu, Teven~Le Scao, Sylvain Gugger, Mariama Drame, Quentin Lhoest, and Alexander~M. Rush.
\newblock Transformers: State-of-the-art natural language processing, July 2020.

\bibitem{su2021roformer}
Jianlin Su, Yu~Lu, Shengfeng Pan, Ahmed Murtadha, Bo~Wen, and Yunfeng Liu.
\newblock Roformer: Enhanced transformer with rotary position embedding.
\newblock {\em arXiv preprint arXiv:2104.09864}, 2021.

\end{thebibliography}
% \printbibliography
\end{document}